\documentclass[letterpaper,10pt,conference]{ieeeconf}
\IEEEoverridecommandlockouts
\usepackage[T1]{fontenc}
\usepackage{times,mathptmx,amsmath,amssymb}
\usepackage{graphicx,booktabs,array,tabularx,microtype,url,balance}
\usepackage{stmaryrd}
\usepackage{bm}
\usepackage{algorithm}
\usepackage{algpseudocode}
\usepackage{xcolor}
\usepackage{setspace}

\AtBeginDocument{\setlength{\parskip}{0pt}}

\newcolumntype{Y}{>{\raggedright\arraybackslash}X}

\newcommand{\R}{\mathbb{R}}

\newcommand{\X}{\mathcal{X}}
\newcommand{\U}{\mathcal{U}}

\newcommand{\N}{{\mathbb{N}}}

\newcommand{\Post}{\textrm{Post}}

\newcommand{\norm}[1]{\lVert #1\rVert}

\newcommand{\nom}{\mathrm{nom}}
\newcommand{\abs}{\mathrm{abs}}

\newcommand{\Obs}{\mathcal{O}}
\newcommand{\Obsafe}{\Obs^{d_{\mathrm{safe}}}}

\newcommand{\Xws}{X_{\mathrm{ws}}}
\newcommand{\Goal}{\mathcal{X}_{\mathrm{goal}}}
\DeclareMathOperator{\dist}{dist}

\DeclareMathOperator*{\argmin}{arg\,min}
\newcommand{\set}[1]{\{#1\}}

\newcommand{\bft}{\textbf}
\newcommand{\des}{\mathrm{des}}
\newcommand{\cert}{\mathrm{cert}}

\usepackage[nolist]{acronym} 

\newacro{ds}[DS]{Dynamical System}
\newacro{lfd}[LfD]{Learning from Demonstration}
\newacro{sos}[SOS]{Sum-Of-Squares}
\newacro{snnds}[S$^2$-NNDS]{Safe and Stable Neural Network Dynamical Systems}
\newacro{pac}[PAC]{Provably Approximately Correct}
\newacro{smt}[SMT]{Satisfiability Modulo Theories}
\newacro{abcds}[ABC-DS]{obstacle Avoidance with Barrier-Certified polynomial
Dynamical Systems}
\newacro{mse}[MSE]{Mean Square Error}
\newacro{lpv}[LPV]{Linear Parameter-Varying}
\newacro{gmm}[GMM]{Gaussian Mixture Models}
\newacro{}[]{}
\newacro{}[]{}
\newacro{}[]{}
\newacro{}[]{}
\newacro{}[]{}
\newacro{}[]{}

\title{Learning from Demonstrations Meets Symbolic Control - A Supervisory Control Architecture for Formal Certification of Learned Motion}
\author{
Ashok Kumar\textsuperscript{1,*},
Mahathi Anand\textsuperscript{2,*},
Angela P. Schoellig\textsuperscript{2},
and Abdalla Swikir\textsuperscript{1}%
\thanks{$^{*}$Ashok Kumar and Mahathi Anand contributed equally.}%
\thanks{$^{1}$Mohamed bin Zayed University of Artificial Intelligence,
Abu Dhabi, UAE.}%
\thanks{$^{2}$Learning Systems and Robotics Lab,
Technical University of Munich, Germany.}%
\thanks{Corresponding author: Ashok Kumar
(email: ashok.kumar@mbzuai.ac.ae).}%
}
\begin{document}
\maketitle
\thispagestyle{empty}\pagestyle{empty}

\begin{abstract}

Dynamical system (DS)-based learning from demonstration generates flexible robot motions, but existing guarantees rely on certificate generation to obtain formal guarantees of safety and goal convergence. We propose a symbolic supervisory control architecture that automatically generates a fail-safe backup plan, certifying safety, reachability, and actuator feasibility directly from the robot's dynamics, independent of the learned DS. A nominal controller tracks the learned DS, while the supervisory controller monitors safety and target convergence, providing non-conservative intervention when the nominal controller violates necessary constraints. Experimental results include simulations and real robot deployment with various 2D and 3D datasets, including handwriting datasets, validating the effectiveness of our approach in providing formal guarantees with learned motion even in complex, dynamic environments.

\end{abstract}

\section{Introduction}

\ac{lfd} is an efficient paradigm to teach robots highly complex tasks directly through demonstrations~\cite{BillardCalinonDilmannSchall2008}. A popular approach is the \ac{ds} method~\cite{BillardMirrazaviFigueroa2022book}, where the demonstrations are encoded via a dynamical system. Here, a vector field governs the velocity of the desired trajectories and issues commands to the low-level controller to track. The key advantage is generalizability: the \ac{ds} is a time-invariant state-feedback policy over the whole domain, it can be evaluated from any state, and can quickly react to runtime perturbations. Nevertheless, additional properties such as stability and safety need to be imposed in order to ensure goal reaching, robustness to perturbations, and collision avoidance. Moreover, the learned vector field is not guaranteed to generate feasible trajectories that satisfy velocity and actuator constraints, as these limits are only imposed implicitly through robot demonstrations, if at all. 

In this paper, we propose a novel approach for safe and robust realization of learned trajectories via symbolic supervisory control. Agnostic to the method used to learn the \ac{ds}, our framework operates at the control layer as a \emph{certified supervisor} that verifies the safety, reachability, and realizability of the resulting motion at runtime. When the nominal tracking controller threatens these guarantees, a symbolic controller intervenes to produce a certified input respecting actuator limits. The resulting motion is guaranteed to reach the goal set within finite time, remain safe against (possibly dynamic) obstacles, and recover from deviations once the environment is clear.
\subsection{Related Work}
\label{subsec:related}

Formal stability and safety guarantees on learned dynamical systems are typically certified through certificate construction --- \emph{Lyapunov functions} for asymptotic stability ~\cite{khalil_nonlinear_2013} and \emph{barrier functions} for safety~\cite{Prajna2004safety}.  A common strategy is to synthesize these certificates concurrently with the \ac{ds}~\cite{seds, figueroa-ds-gmm, lpv, abc, s2nnds}. The synthesis strategy, and subsequently, the optimization tool used, vastly depends on the characterization of the \ac{ds}. For instance,~\cite{seds, figueroa-ds-gmm, lpv} learn a \ac{lpv} dynamical system, and use conservative quadratic Lyapunov candidates for certifying stability. On the other hand,~\cite{abc} learn polynomial dynamical systems along with polynomial certificates using sum-of-squares~\cite{ParilloSOS} tools, while~\cite{s2nnds} propose neural network characterizations for the system and certificates. 

Independent of parameterization, certificate search suffers from a fundamental trade-off between the expressiveness of the learned dynamics and the tractability of the underlying optimization. An alternative is to modify the learned stable dynamics at runtime~\cite{khansari-zadeh_dynamical_2012, huber_avoidance_2019}, eliminating certificate optimization for collision avoidance, or to pass the generated velocity through a safety filter via a quadratic program with hand-crafted control Lyapunov and barrier functions~\cite{nawaz}. Our approach departs from these in three respects. First, rather than modifying the learned dynamics, we operate one layer below it, designing control inputs that track the \ac{ds} closely and intervening only when obstacles block the desired path. Additionally,  physical realizability is guaranteed through velocity and actuator constraint satisfaction. Second, while Lyapunov-based approaches guarantee only asymptotic convergence, we guarantee reachability of a neighborhood of the goal within finite time. Finally, our synthesis does not require any certificate functions, only finite state abstractions of the model.

Our work also borrows from symbolic control, where finite approximations of control systems inform controller synthesis through simple fixed point computations~\cite{tabuada2009verification,belta_symbolic_2007}. This has been applied primarily in the context of motion planning for mobile robots and autonomous vehicles~\cite{fainekos_temporal_2009,symbolic_vehicle}.
While~\cite{csymplan} addresses motion planning for manipulators via symbolic control, we utilize symbolic control strategies for safely modifying controllers obtained via \ac{lfd}.    

\subsection{Contributions}

\begin{figure*}[t]
\setlength{\belowcaptionskip}{-6pt}
\centering
\includegraphics[width=1.0\textwidth]{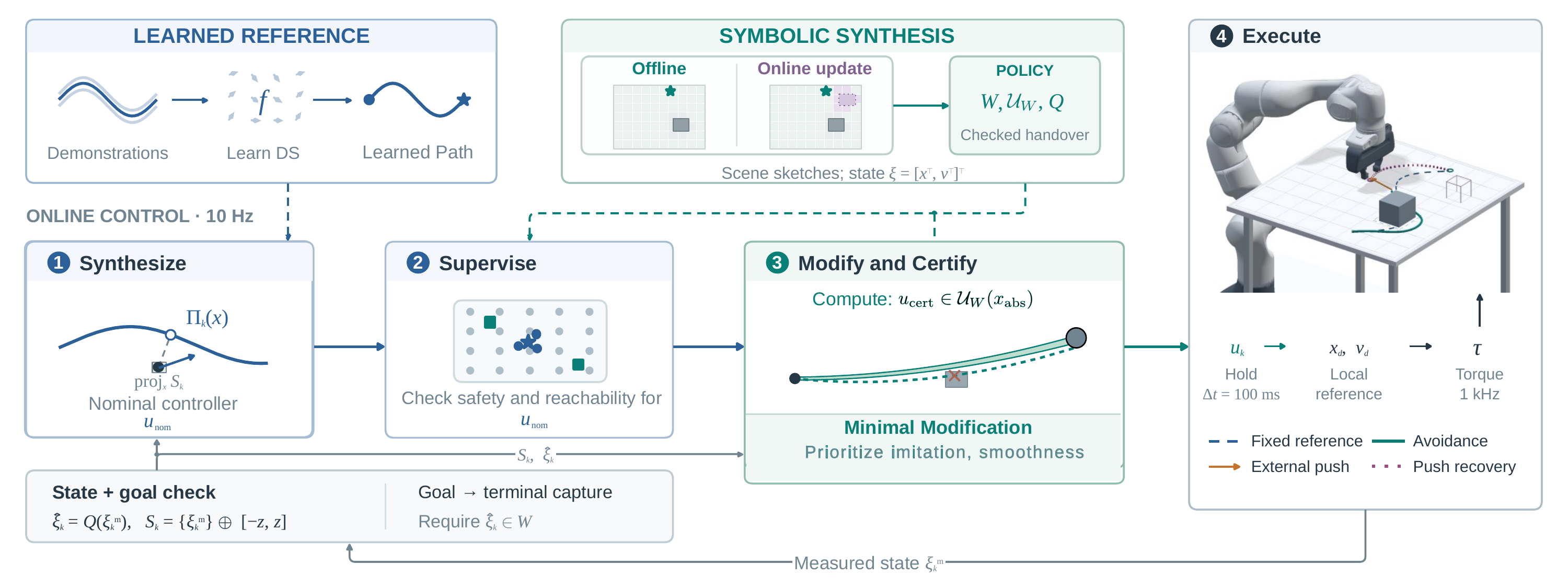}
\caption{\centering The proposed pipeline follows the principle \textit{synthesize, supervise, and certify}. Starting with the learned DS and a symbolic certified controller, our methodology synthesizes nominal controllers to follow the learned paths, supervises and checks for collision and reachability to the goal, and modifies the controller slightly to obtain certified inputs before executing them on the robot. }
\label{fig:overview}
\end{figure*}

The contributions of our paper are as follows:
\begin{enumerate}
\item Starting with a~\ac{ds} learned from demonstrations with a preferred method, we propose a novel \emph{supervisory architecture} that functions in the control layer and (i) monitors the safety of the reference paths generated by the~\ac{ds}, and (ii) modifies it so the robot can still reach a small neighborhood of the goal, while ensuring collision avoidance against dynamic obstacles. The designed methodology avoids computation of certificate functions while retaining formal guarantees of reachability and safety under standard assumptions on the real world (c.f. Section~\ref{subsec:supervisory}).  
\item Our architecture relies on finite state abstractions on the manipulator's operational space dynamics through sample-perturbed-system representations. Based on the discretized system dynamics and regions of interest (i.e, goals and obstacles), a symbolic controller is constructed as a fail-safe certified backup plan. Synthesis is performed either \emph{offline}, suited for static obstacles, or \emph{online} at low frequencies for dynamic obstacles.
\item At runtime, a nominal tracking controller is designed to faithfully follow the paths generated by the~\ac{ds}, and ensures convergence to the desired paths in the presence of deviations. A certified control input, computed to provide less-conservative modifications based on a ranking scheme, overrides the nominal control action when it is rendered to be unsafe or not goal-reaching. A full overview of our approach is presented in Figure~\ref{fig:overview}.
\item We validate our approach through several simulation experiments (MuJoCo) for learning handwriting motion with static and obstacles, and particularly compare the performance of approach against the closely related-work~\cite{nawaz}. The results show that our approach is competitive with SoA and excels with high task success rates and formal guarantees. Additionally, our robot experiments on dynamic $3D$ environments emphasize the flexibility of our approach while demonstrating that formal guarantees hold even in the real world. 
\end{enumerate}

\section{Preliminaries} \label{sec:prelim}
\subsection{DS-based Learning from Demonstrations}
\label{subsec:ds}
Suppose that we are given a set of $N$ robot demonstrations, $\mathcal{D} = \big\{x_{ij}, \dot x_{ij} \big \}_{i=1, j=1}^{N,M}$, where $x \in \Xws \subseteq \R^d$ represent Cartesian positions of the robot in its workspace $\Xws$, and $\dot x \in \R^d$ represents the Cartesian velocities, such that each demonstration consists of time-discretized trajectories $\big ( x_i(t_1),  x_i(t_2), \ldots,  x_i(t_M) \big)$ and $\big (\dot{ x}_i(t_1), \dot{ x}_i(t_2),  \ldots, \dot{ x}_i(t_M) \big)$,  $t_0 = 0$, $t_M > 0$, $i \in \{1,\ldots,N\}$. The \ac{lfd} framework aims to learn a \ac{ds} described by
\begin{equation} \label{eq:ds}
\dot{x}(t) = f(x(t)),
\end{equation}
that imitates the demonstrations closely, while reaching a goal region $\Goal$ while staying safe and avoiding a set of obstacles $\Obs$. The standard approach is to learn a \ac{ds} by simply minimizing the \ac{mse}:
\begin{equation}
\min_f \frac{1}{MN} \sum_{i=1}^N \sum_{i=1}^M \|\dot x_{ij} - f(x_{ij}) \|^2.
\label{eq:mse}
\end{equation}
Solving~\eqref{eq:mse} is generally computationally intractable since it is an infinite-dimensional optimization problem. In practice, function parameterizations for $f$ enable scalable learning. Examples include \ac{lpv} dynamics using \ac{gmm}~\cite{seds, lpv}, polynomial dynamics~\cite{abc}, or neural network parameterizations~\cite{nawaz, s2nnds}. The learned $f$ then provides desired position and velocity commands, which are then tracked by a low-level robot controller under nominal circumstances in the absence of disturbances or obstacles. 

However, since $f$ is learned using finite amount of data, and safety is only implicitly represented through demonstrations, the learned \ac{ds} is neither guaranteed to stay away from obstacles, nor reach the goal in the presence of perturbations. To achieve safety and stability guarantees, certificate functions like barrier or Lyapunov functions, are typically employed. Also, note that \ac{ds}-based formulation is not guaranteed to be physically feasible - indeed, in some cases, a poorly learned $f$ may take the trajectory outside of the workspace $\Xws$, and may even violate the velocity constraints or actuator limits. 

In our work, we take an alternate view point. Rather than relying on carefully crafted certificate functions, we use a supervisory control architecture to modify the nominal control input in real time when the tracking input violates safety or the constraints imposed on the robot. Our proposed approach is agnostic to the learning method used, and relies on a formally certified symbolic synthesis architecture developed on an acceleration-controlled reference model. Some preliminary background is presented below.

\subsection{Sample Perturbed Systems}
\label{subsec:sample-perturbed}

We describe a sample perturbed system by a tuple $S=(\X,\U,\mathcal{F})$, where $\X\subseteq\R^n$ is the state set, $U\subseteq\R^m$ is the input set, and $\mathcal{F}: \X\times \U\rightrightarrows X$ is a set-valued transition map. We consider a continuous-time 
inclusion
\begin{equation}
    \dot\xi\in f(\xi,u)\oplus\llbracket-w,w\rrbracket,
    \label{eq:symbolic_dyn}
\end{equation}
where $\xi \in \X$, $u\in \U$, and $w\in\R^n_{\ge0}$ is the disturbance set
\begin{equation*}
    \llbracket-w,w\rrbracket=[-w_1,w_1]\times\cdots\times[-w_n,w_n].
\end{equation*}
Measurement uncertainty is represented by \eqref{eq:symbolic_meas}:
\begin{equation}
    P(x)=\xi \oplus\llbracket-z,z\rrbracket,
    \label{eq:symbolic_meas}
\end{equation}
with $z\in\R^n_{\ge0}$. We represent the $\Delta t$-sampled behavior of the above system with a $\xi^{+} \in \hat f(\xi, u)$ as a solution of~\eqref{eq:symbolic_dyn} under a held input $u \in \U$ within the time interval $\Delta t$. This induces a discrete-time behavior of $S$ is an infinite sequence $(\bm{\xi}, \bm{u}) = (\xi_0,u_0,\xi_1,u_1,\ldots)$, where $\xi$ collects the sequence of states such that $\xi_{k+1}\in \mathcal{F}(\xi_k,u_k)$ for all $k\in\N$.

\subsection{Acceleration Model based on Operational Space Dynamics}
\label{subsec:os-dyn}

Consider an $n_q$-DOF manipulator with joint configuration $q \in \R^{n_q}$ and the rigid body dynamics:
\begin{equation}
    M(q)\ddot q+C(q,\dot q)\dot q+g(q)=\tau+\tau_{\mathrm{ext}}.
    \label{eq:manip_joint_dyn}
\end{equation}
Here, $M(q)$ is the inertia matrix, $C(q,\dot q)\dot q$ contains Coriolis and centrifugal terms, $g(q)$ is gravity, $\tau$ is the commanded joint torque, and $\tau_{\mathrm{ext}}$ is external torque. Let the end effector's translational position be $x \in \R^d$, where $x = h(q)$. Correspondingly, the velocity is $\dot x = J(q) \dot q$ where $J(q) = \frac{\partial h}{\partial q}$. Then, we have from~\cite{khatib_os} the operational space robot dynamics:
\begin{equation}
    \Lambda(q)\ddot x+\eta(q,\dot q)=F+F_{\mathrm{ext}},
    \label{eq:opspace_dyn}
\end{equation}
where $\Lambda(q)=(J M^{-1}J^\top)^{-1}$ is the operational-space inertia, $\eta(q,\dot q)$ groups task-space bias terms, $F\in\R^d$ is the commanded task-space wrench for the translational task, and $F_{\mathrm{ext}}$ collects external task-space effects. Now, we choose
\begin{equation}
    F=\eta(q,\dot q)+\Lambda(q)u,
    \label{eq:task-command}
\end{equation}
where $u\in\R^d$ is a virtual input as desired end effector acceleration. Substituting \eqref{eq:task-command} into \eqref{eq:opspace_dyn} gives
\begin{equation}
    \ddot x=u+\Lambda(q)^{-1}F_{\mathrm{ext}}.
    \label{eq:ddx_virtual}
\end{equation}
By bounding the residual term in~\eqref{eq:ddx_virtual}, together with discretization, measurement as well as torque-realization errors as a bounded additive disturbance $\llbracket -w, w \rrbracket$, and using $\xi = [ x^\top, v^\top]^\top$ with $v = \dot x$, we get
\begin{equation}
    \dot\xi=
    \begin{bmatrix}
0&I_d\\ 0&0
    \end{bmatrix}\xi+
    \begin{bmatrix}
0\\ I_d
    \end{bmatrix}u
\oplus\llbracket-w,w\rrbracket.
    \label{eq:sp_mani_sys}
\end{equation}
Note that the input set $\U$ is compact and is selected to represent actuation limits after torque realization, while $\X \subseteq \R^{2d}$ is the robot state set.

\subsection{Symbolic Controller Synthesis}
\label{subsec:symbolic}

In this work, we are concerned with ensuring that the learned trajectories, and their resultant robot realization, reaches the desired goal region while remaining safe. To this end, we define a reach-and-avoid specification for the sample-perturbed system $S$ as follows:
\begin{align} \label{eq:ra}
    \Sigma_{\mathrm{RA}} := \{\bm \xi \mid \xi_0 \in \X_{\mathrm{0}} & \implies \exists T \text{ s.t. } \xi_T \in \X_{\mathrm{goal}}; \nonumber \\
    &\xi_k \in \X_{\mathrm{safe}}, \forall k \in k \in \{0, \ldots, T\} \},
\end{align}
where $\X_{\mathrm{safe}}$ and $\X_{\mathrm{goal}}$ are the safe and goal regions, respctively. Note that these specifications are described in discrete-time following the discrete behavior of the sampled perturbed system~\eqref{eq:symbolic_dyn}. 

To obtain suitable inputs $u$ for the sampled perturbed system against the specification $\Sigma_{RA}$, we employ symbolic control, introduced for robot manipulators in~\cite{csymplan}. It constructs a finite abstraction $S_{\mathrm{abs}} = (\X_{\mathrm{abs}}, \U_{\mathrm{abs}}, F_{\mathrm{abs}})$ that overapproximates the system $S$ via a quantizer function $Q: \X \rightarrow \X_{\mathrm{abs}}$, resulting in a graph-based relation 
\begin{equation*}
    R_Q=\set{(\xi,\hat \xi)\in \X\times  \X_{\mathrm{abs}}\mid \hat \xi=Q(\xi)},
\end{equation*}
Assuming that $\U_{\mathrm{abs}} \subseteq \U$ for admissibility, if we have for every $(\xi,\hat \xi)\in R_Q$; 
\begin{enumerate}
\item $\U_{\mathrm{abs}}(\hat \xi)\subseteq \U(\xi)$; and
\item for all $u\in \U_{\mathrm{abs}}(\hat \xi)$, $Q(\mathcal{F}(\xi,u))\subseteq \mathcal{F}_{\mathrm{abs}}(\hat \xi,u)$, where $Q(\mathcal{F}(\xi,u))=\set{Q(x^+)\mid \xi^+\in \mathcal{F}(\xi,u)}$,
\end{enumerate}
then, we say that $R_Q$ is a feedback refinement relation, and consequently, any safe control input $u \in \U_{\mathrm{abs}}$ desgined for $S_{\mathrm{abs}}$ may be directly utilized as a control input $u$ for the system $S$. If the controller is such that $S_{\mathrm{abs}}$ satisfies $\Sigma_{RA}$, then the guarantees are also maintained for $S$~\cite{reissig_feedback_2014}. 

\section{Proposed Method}

This section presents the main results of our paper. Given a learned \ac{ds}, the goal is to design a supervisory controller that enables the robot to navigate safely according to the reference paths generated by the \ac{ds} in an obstacle-rich real world. In other words, the designed controller should (i) track the desired paths and velocities obtained via~\eqref{eq:ds} where possible, (ii) ensure that the robot reaches close to the desired goal, even when $f(\cdot)$ does not have a fixed point at the goal, (iii) safely modify the reference paths obtained from the DS and navigate around the obstacle when needed, and finally (iv) recover the original path when it is clear from obstacles. Through our method, we aim to achieve formal reach-and-avoid guarantees even with (\ac{ds}) learning-in-the-loop. 

\subsection{Reach-and-Avoid Characterization}

Suppose that there exist obstacles in the workspace $\Xws$ and $\Obs$ represents their union. We define the inflated obstacle set in order to ensure sufficient clearance from the obstacles:
\begin{equation}
\Obsafe := \big \{x \in \Xws \mid \mathrm{dist}(x, \Obs) \leq d_{\mathrm{safe}} \big \}
\end{equation}
\begin{equation*}
X^x_{\mathrm{safe}}=X_{\mathrm{ws}}\setminus\Obs^{d_{\mathrm{safe}}}.
\end{equation*}
Lifting this set to the full robot state gives
\begin{equation}
    \X_{\mathrm{safe}}=\set{\xi=[x^\top,v^\top]^\top\in\R^{2d}\mid x\in X^x_{\mathrm{safe}},\ v\in V},
    \label{eq:safe_set}
\end{equation}
where $V \subset \R^d$ is a compact velocity set describing the velocity limits for the robot. Similarly, for a desired target position $x_T \in X^x_{\mathrm{safe}}$ and a tolerance $d_{\mathrm{tol}}$, we define 
\begin{equation*}
    X^x_{\mathrm{goal}}=\set{x\in X_{\mathrm{ws}}\mid \dist(x,x_T)\le d_{\mathrm{tol}}},
\end{equation*}
over the robot position, and since the robot should come at rest after reaching $X^x_{\mathrm{goal}}$, we have
\begin{equation}
    \X_{\mathrm{goal}}=\set{\xi=[x^\top,v^\top]^\top\in\R^{2d}\mid x\in X^x_{\mathrm{goal}},\ \norm{v}_{\infty}\le v_{\mathrm{tol}}}.
    \label{eq:goal_set}
\end{equation}

\subsection{Nominal Controller Extraction}
\label{subsec:u_nom}
Before we describe the supervisory control architecture, we first extrapolate a nominal controller from the reference trajectory generated by the \ac{ds}~\eqref{eq:ds}. Typically, passive impedence control is used for this purpose~\cite{KronanderBillard2016PassiveInteractionControl}, but is not directly applicable for our feedback linearizable, sample-perturbed system representation. From a given initial point $x_0$, the integration $x_t = \int_{0}^T f(x(t)) \mathrm{d}t$ gives the time-integrated open-loop path $\mathrm{x}_\des = x_0, \ldots, x_T$. However, the time dependence results in low robustness, as any deviation from the path (e.g., due to to safety concerns) may result in poor tracking behavior. Therefore, we seek to re-parameterize the path as a function of the arc-length rather than time. Specifically, define the arc-length as
\begin{equation} \label{eq:arc_length}
s(T) = \int_{0}^T \|f(x(t))\| \mathrm{d}t,
\end{equation}
where $s(t)$ can be interpreted as the distance traveled along the path $\mathrm{x}_\des$ upto time $t$. Correspondingly, define the curve $\Gamma[s]$ parameterized by the length $s$ as $\Gamma[s] := x(t(s))$, so that sampling $\Gamma$ at arc-lengths $s_i$ yields waypoints that are evenly spaced over the desired path $\mathrm{x}_\des$. Note that $\mathrm{x}_\des$ and thus $\Gamma$ are not guaranteed to be goal-reaching as the vector field $f(\cdot)$ is learned without stability guarantees. Additionally, we store the desired velocity profiles $\bar v_i = f(r_i)$ at $r_i = \Gamma(s_i)$, and linearly interpolate between neighboring samples to obtain
\begin{equation}
 \bar v(s)=(1-\alpha)\bar v_i+\alpha\bar v_{i+1},\qquad
 \alpha=\frac{s-s_i}{s_{i+1}-s_i},\quad s\in[s_i,s_{i+1}].
 \label{eq:velprofile}
\end{equation}
As a result, the reference signals $\Gamma(s)$ and the corresponding velocity profiles $\bar v(s)$ define spatial reference paths for the robot positions and velocities respectively. 

At each control step $k$, the robot maps its position with respect to the reference paths, in order to estimate its own progress. To do so, given a robot position $x \in \R^d$, we employ the nearest-point projection $\Pi_k(x) = \Gamma(s_k(x))$, where $s_k(x) = \arg\min_{s\in I_k} \norm{x-\Gamma(s)}_2^2$ is the arc length of the closest point in $\Gamma$ over an eligible time interval $s \in I_k$. Note that we do not consider previously measured progress and instead trigger fresh localization at every step since any safety interventions or perturbations may invalidate the progress made thus far. 

We also measure the safety of the projected state $\Pi_k(x)$, selected among eligible return targets with a clear connection, via a Boolean variable $\chi_k \in \{0, 1\}$ defined as
\begin{equation*}
\chi_k = \begin{cases}
1 & \Pi_k(x) \in X^x_{\mathrm{safe}}, \\
0 & \text{otherwise}.
\end{cases}
\end{equation*}
$\chi$ indicates whether it is permissible for the robot to return to the path $\Gamma$, and triggers a path-return preference to ensure that any deviated paths converge to the desired path. In addition, we define an optional speed envelope $\sigma(s)$ that upper-bounds the magnitude of the preferred velocity along $\Gamma$, and allows the robot to move at the demonstrated speed when the path is sufficiently smooth, and slows it down in the presence of non-smooth curves and bends. Consequently, we define a function $l(\bar v(s), \sigma(s))$ that caps $\bar v(s) = \sigma(s)$ if $\norm{\bar v(s)}_2 >= \sigma(s)$. 
This enables the design of a preferred velocity profile:
\begin{equation}\label{vel:profile}
h_k(x) :=
\begin{cases}
l\!\left(f_\theta(\Pi_k(x)),\,\sigma(s)\right)
-k_\perp\!\left(x-\Pi_k(x)\right),
& \chi_k=1,\\[1mm]
f_\theta(x),
& \chi_k=0.
\end{cases}
\end{equation}
where $k_\perp > 0$ is a high return gain. $h_k(x)$ is constructed such that it encourages the robot to return to the desired path in the absence of any obstruction. We obtain the acceleration as
\begin{equation} \label{eq:acc_nom}
  \bar u= \frac{\partial h_k(x)}{\partial x}v
                  +K\big(h_k(x)-v\big),
\end{equation}
where $v$ is the robot velocity at time $k$ and $K$ is feedback gain on the velocity tracking error. Finally, to ensure path following and minimal acceleration variation, we modify $\bar u$ with the following cost function to obtain the nominal controller:
\begin{align} 
u_\mathrm{nom} = \argmin_{u \in \mathcal{U}} &\frac{\chi_k w_p}{M}\sum_{j=1}^M
                       \norm{x_j-\Pi_k(x_j)}_2^2 
+ \frac{w_a}{M}\norm{\bar u-u_{nom}}_2^2 \nonumber \\             
& +w_\Delta\norm{u-u_{k-1}}_2^2, \label{eq:nom}
\end{align}
where $w_p, w_a, w_\Delta > 0$ are appropriate cost function weights. Notice that the first term in~\eqref{eq:nom} prioritizes inputs that render the predicted path close to the desired path obtained via the DS~\eqref{eq:ds}, and is inactive when the path is already along the desired one (i.e., $\chi = 0)$. The second term ensures minimal deviations from the nominal controller, and the last term guides smooth motions by minimizing acceleration deviations. 

\subsection{Supervisory Control Methodology}
\label{subsec:supervisory}

The nominal controller $u_\nom$ constructed above allows for tracking desired position and velocity paths, and ensuring that the original path is recovered in presence of any deviations. Nevertheless, as written earlier, it doesn't enable safety nor goal reaching unless the learned \ac{ds} inherently possesses such properties. To this end, we propose a supervisory control architecture based on symbolic controller synthesis, that runs on backend to (i) verify at a given time instant whether $u_\nom$ indeed generates a safe trajectory, and (ii) provides a non-restrictive control input $u_\cert$ as a backup controller that drives the robot to safety, while allowing recovery to the goal along the desired path even in the presence of disturbances. First, we present an offline architecture implementation, shown in Algorithm~\ref{alg:offline_synthesis} and adapted from CSymPlan~\cite{csymplan} to generate a one-time synthesis pipeline to search for a finite set of modified controller inputs. This algorithm runs only once before the overall task execution and handles static obstacles and runtime perturbations. Then, we will also present an online version that can operate at runtime and tackles dynamic obstacles.  

For a fixed workspace, obstacle set and goal set (which are generally determined from the demonstrations), we first construct a finite abstraction $S_\mathrm{abs}$ of~\eqref{eq:sp_mani_sys}. Specifically, the continuous state $\X$, safe $\X_{\mathrm{safe}}$ and goal sets $\X_{\mathrm{goal}}$ are discretized into uniform grids $\X_{abs}$, $\X_{\mathrm{safe}, \mathrm{abs}}$, and $\X_{\mathrm{goal}, \mathrm{abs}}$, respectively, and the acceleration input set $\U$ is discretized into $\U_{\mathrm{abs}}$. Consequently, for every abstract state-input pair, $\mathcal{F}_\mathrm{abs}$ is also determined by over-approximating all reachable sampled successors under the disturbance and measurement bounds $\llbracket -w, w \rrbracket$.
A fixed point computation on $S_\mathrm{abs}$ yields a winning set, i.e., a set of quantized states in $W \subseteq \X_\mathrm{abs}$ that contains all the states that remain safe while reaching the desired goal, and a corresponding set of winning control inputs $\U_W(\xi_\mathrm{abs})$, for all $\xi_\mathrm{abs} \in W$ such that we have $\mathcal{F}_{\mathrm{abs}}(\xi_\mathrm{abs}, u_\mathrm{abs}) \in W, \forall u_\mathrm{abs} \in \U_W(\xi_\mathrm{abs})$. The resulting controller is stored as a binary decision diagram or a lookup table for use during runtime execution. 

\begin{algorithm}[t]
\caption{Offline synthesis algorithm}
\label{alg:offline_synthesis}
\begin{spacing}{0.95}
\footnotesize
\begin{algorithmic}[1]
\Require Workspace bounds $X_{\mathrm{ws}}$, obstacles $\Obs$, safety margin $d_{\mathrm{safe}}$, velocity bounds $V$
\Require Sampling time $s$, disturbance and measurement bounds $w,z$, grid parameters, input grid $U_{\mathrm{abs}}\subseteq\U$
\Require Task specification: initial set $X_0$, goal $X_{\mathrm{goal}}$
\State Compute $\X_{\mathrm{safe}}$ using \eqref{eq:safe_set}
\State Discretize $\X=X_{\mathrm{ws}}\times V$ into uniform cells to obtain $\X_{\mathrm{abs}}$
\State Compute $\mathcal{F}_{\mathrm{abs}}$ by over-approximating reachable sets of \eqref{eq:sp_mani_sys}
\State Build the quantizer $Q:\X\to \X_{\mathrm{abs}}$
\State Construct abstract sets $\X_{0,\mathrm{abs}}$, $\X_{\mathrm{safe},\mathrm{abs}}$, and $\X_{\mathrm{goal},\mathrm{abs}}$
\State Synthesize a winning set $W$ and corresponding admissible input map $U_W(x_{\mathrm{abs}})$ for all $x_{\mathrm{abs}}\in\X_{\mathrm{abs}}$ by fixed-point computation
\State Export $U_W(\hat\xi), \forall \hat \xi \in \X_{\abs}$ as a BDD or lookup table\end{algorithmic}
\end{spacing}
\end{algorithm}

At runtime, the offline synthesized symbolic policy will act as a supervisory controller that provides overrides the nominal controller when it fails to keep the robot from satisfying the necessary reachability and safety requirements. In particular, for a given robot state $\xi = [x^\top, v^\top]$, we compute $u_\nom$ as described in Section~\ref{subsec:u_nom}, and predict the resulting task-space path over the current hold interval $[0, \Delta t]$ via the sample perturbed model~\eqref{eq:sp_mani_sys} as
\begin{equation} \label{eq:samp-unom}
x(t) = x + t v + \frac{1}{2} t^2u_\nom, \ v(t) = v + t u_\nom, \ t \in [0, \Delta t].
\end{equation}
Overapproximating the predicted trajectory together with the disturbance bounds $\llbracket -w, w \rrbracket$ gives a set of states $\mathcal{T}(S, u_\mathrm{nom})$, which are then quantized according to the map $Q$ to obtain a set of successor cells that we denote using $\Post(S, u_\nom)$. If $\emptyset \neq \Post(S, u_\nom) \subseteq W$, then the entire predicted trajectory remains within the certified safe set $\X_{\mathrm{safe}}$, and the trajectory is guaranteed to reach the goal set $\X_{\mathrm{goal}}$. Then, $u_{\mathrm{cert}} = u_{\mathrm{nom}}$ is applied to the robot. On the other hand, if $\Post(S, u_\nom) = \emptyset$ or $\Post(S, u_\nom) \not\subseteq W$, then a backup control input is required. Specifically, at the given quantized robot state $\xi_\mathrm{abs} = Q(\xi)$, the lookup table provides a set of safe, feasible control actions $\U_W(\xi_\abs)$. Among the candidates $u \in \U_W(\xi_\abs)$, we compute the minimally restrictive certified control input by modifying~\eqref{eq:nom} as
\begin{align} 
u_\mathrm{cert} \hspace{-0.1em}= & \hspace{-0.2em} \argmin_{u \in \U_W(\xi_{j,\mathrm{abs}})} \hspace{-0.1em}  \frac{\chi_k w_p}{M}\sum_{j=1}^M
                       \norm{x_j-\Pi_k(x_j)}_2^2 \nonumber \\
 & +\frac{w_v}{M}\sum_{j=1}^M
            \norm{v_j-h_k(x_j)}_2^2  
+ w_a \norm{u-u_{nom}}_2^2           \nonumber \\ 
&+w_\Delta\norm{u-u_{k-1}}_2^2, \label{eq:cert}
\end{align}
where $x_j=x + t_jv_m+\tfrac12t_j^2 u$ and
$v_j=v+t_ju $, with $t_j=\frac{j\Delta t}{M}$, are the $j^{th}$ predicted position and velocity of a trajectory of length $M$, $u_{k-1}$ is the last held control input, $w_v > 0$ and the additional term (second) ensures minimal deviation with respect to the computed velocity profile~\eqref{eq:velprofile}.
Algorithm~\ref{alg:online_supervisor} describes our proposed online supervisory control mechanism.



While the above approach is catered only to static obstacles, it is also possible to adapt Algorithm~\ref{alg:offline_synthesis} for implementation in an online manner, which allows for safe supervisory control against dynamic obstacles. Specifically, the online implementation moves the symbolic synthesis call into the robot runtime, and an online server continuously updates the symbolic grid as well as the input grid, $\Post(\cdot)$ operator, and the growth bound arising from the sampled-perturbed model $S$ in~\eqref{eq:sp_mani_sys}.  A separate robot process sends the current task data to the server, updates the safe and goal sets based on current robot state $\xi$,  requests a finite-horizon reach-avoid controller, receives one or more admissible symbolic accelerations, which are then evaluated and executed via Algorithm~\ref{alg:online_supervisor}. We refrain from presenting a detailed discussion on the online implementation, but refer the readers to~\cite[Algorithm 1, 2]{csymplan} for a thorough description and analysis.

\begin{algorithm}[t]
\caption{Online supervisory control at decision $k$}
\label{alg:online_supervisor}
\begin{spacing}{0.95}
\footnotesize
\begin{algorithmic}[1]
\Require Offline policy $U_W(\cdot)$, quantizer $Q$, winning set $W$, hold interval $\Delta t$, horizon length $M$, weights $w_p,w_v,w_\Delta$, previous input $u_{k-1}$
\State Measure state $\xi=[x^\top,v^\top]^\top$
\State Compute $u_{\mathrm{nom}}$ as in Section~\ref{subsec:u_nom}
\State Predict trajectory $x(t),v(t)$, $t\in[0,\Delta t]$ via \eqref{eq:samp-unom}
\State Enclose predicted trajectory with disturbance bounds to obtain $\mathcal T(S,u_{\mathrm{nom}})$
\State Quantize: $\mathrm{Post}(S,u_{\mathrm{nom}}) \gets Q(\mathcal T(S,u_{\mathrm{nom}}))$
\If{$\varnothing \neq \mathrm{Post}(S,u_{\mathrm{nom}}) \subseteq W$}
    \State $u_k \gets u_{\mathrm{nom}}$ \Comment{Nominal input certified safe and goal-reaching}
\Else
    \State $x_{\mathrm{abs}} \gets Q(x)$
    \State Retrieve candidate set $U_W(x_{\mathrm{abs}})$ from lookup table
    \State $u_k \gets u_{\mathrm{cert}}$
    \Comment{solution to~\eqref{eq:cert}}
\EndIf
\State Realize $u_k$ to joint torque $\tau_k$ via operational-space torque control \Comment{solution to~\eqref{eq:torque_realization}}
\State Compute $\tau_k$ via $u_k$ and apply to the robot; set previous input $u_{k-1}\gets u_k$
\end{algorithmic}
\end{spacing}
\end{algorithm}

\noindent \bft{Runtime Execution.} $u_\mathrm{cert}$ is executed at high frequencies, while Algorithm~\ref{alg:online_supervisor} is called at a lower frequency. At the low decision rate, the supervisor selects and buffers a held input $u_\mathrm{cert}$ and applies it in a sample-and-hold manner until the next safe input is available. The control input $u_\mathrm{cert}$ is executed via
\begin{equation}
F \;=\; \eta(q,\dot q) \;+\; \Lambda(q)\,u_{\mathrm{cert}},
\label{eq:Total_force}
\end{equation}
and the joint torques are obtained through
\begin{equation}
\tau \;=\; J(q)^\top F \;+\; \tau_{0},
\label{eq:torque_realization}
\end{equation}
where $\tau_{0,k}$ is used for secondary objectives operating on the robot's null space.

\noindent \bft{Formal Certification. } Guarantees provided by our algorithm depend on the soundness of the system abstraction. In particular, the a safe, goal-reaching discrete input $u \in \U_{\mathrm{abs}}$ designed over the abstract system $S_\mathrm{abs}$ is guaranteed to be safe and goal-reaching for $S$ if: (i) The abstract transition relation $\mathcal{F}_{\mathrm{abs}}$ over-approximates all sampled perturbed task-space successors of \eqref{eq:sp_mani_sys}, (ii) the abstract safe and goal sets are sound over-approximations of the continuous sets \eqref{eq:safe_set} and \eqref{eq:goal_set}, (iii) the difference between the intended virtual acceleration $u_\mathrm{cert}$ and the acceleration realized by the robot remains within the disturbance bound $w$ used during synthesis, (iv) the workspace and obstacle model match the physical environment after the configured safety inflation, and (v) the margin $d_{\mathrm{safe}}$ covers the end-effector and tool geometry, since the safe set is stated for the translational end-effector position. This result directly follows from the construction of a feedback refinement relation in Section~\ref{subsec:symbolic}~\cite{reissig_feedback_2014}. These guarantees also hold the online symbolic controller synthesis case, but on the additional conditions that (i) the online server always returns a feasible winning set, in the absence of which uncertified $u_\mathrm{nom}$ may be applied without guarantees, and (ii) the environmental changes that invalidate the last computed $u_\mathrm{cert}$ are no more faster than than the (low) frequency at which the online server operators.

\section{Experimental Evaluation}

The effectiveness of our approach is tested both in simulation and real-world scenarios. The results are organized in three different categories:
\begin{enumerate}
\item The performance of our algorithm is illustrated on a representative subset of 2D LASA handwriting datasets\footnote{Publicly available at \url{https://bitbucket.org/khansari/lasahandwritingdataset}} in MuJoCo simulation environments. Our simulations include \ac{ds} learned from different methods including \ac{lpv}-DS~\cite{lpv}, polynomial DS (PolyDS)~\cite{abc}, and neural network DS~\cite{s2nnds}, and provide a qualitative illustration of the supervisory architecture's behavior in the presence of obstacles as well as robot perturbations. 
\item We provide a detailed quantitative analysis and benchmarking comparisons of supervisory control methodology against CLF-CBF safety filtering~\cite{nawaz} for the case of handwriting datasets, and particularly isolate the performance across nominal tracking performance as well as in the presence of robot perturbations and static obstacles. 
\item Finally, we demonstrate the applicability of our approach in the real-world through a 3D dataset, deployed on a Franka FR3 robot.  
\end{enumerate}

\begin{figure*}
\setlength{\belowcaptionskip}{-6pt}
    \centering
    \includegraphics[width=0.9\linewidth]{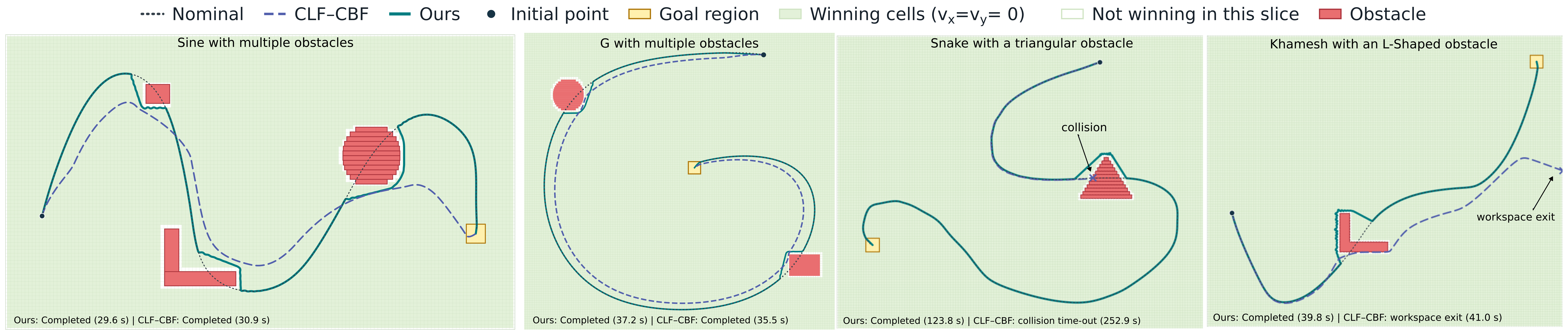}
    \caption{\centering Simulations showing the performance of our supervisory control method vs CLF-CBF~\cite{nawaz} for four representative datasets: sine, G, snake,
    and khamesh, with different obstacle configurations.}
    \label{fig:static}
\end{figure*}

\noindent \textbf{MuJoCo Simulations with 2D LASA Datasets.} First, we consider the LASA handwriting datasets, a standard for benchmarking in \ac{lfd} literature. The datasets are normalized around the mean, and three different \ac{ds} models are trained, corresponding to LPV, polynomial, and NN dynamics. All demonstrations are designed to reach a pre-specified goal point. We consider the Franka FR3 robot simulation environment on MuJoCo, with workspace adapted to the handwriting shapes. The initial and goal region of the trajectories are constrained to be a balls of radius $10$mm and $40$mm around the average initial position and goal of the demonstrations, respectively. The environments are designed to include (possibly many) circular, and box-like obstacles. In addition, velocity constraints $v \in [-0.4, 0.4]$m/s and $u \in [-2,2]$m/s$^2$ are additionally introduced. 
First, we illustrate the \textit{offline} case,  where supervisory policies from Section~\ref{subsec:supervisory} are pre-computed over a fixed workspace prior to task execution through SCOTS~\cite{rungger_scots_2016} using finite abstraction with (possibly) static obstacles in the environment, and exports the controller as a lookup table. Then, at runtime, the robot state is extracted through quantization and a certified control input is computed via Algorithm~\ref{alg:online_supervisor} at $10Hz$, realized on the simulated robot at $1kHz$ through operational-space torque control. A representative simulation is shown in Figure~\ref{fig:static}. The trajectories obtained through our supervisory control method carefully track the nominal trajectories learned by the \ac{ds} until the path is no longer safe; and with a small, non-conservative modification around the obstacle, it returns to the path once the obstacles are cleared. In comparison, CLF-CBF based tracking control may cause QP infeasibility issues leading to collisions\footnote{We add CBF slack constraints~\cite{nawaz} to prevent QP infeasibility at runtime}, or conservative path modifications that cannot converge to the original, desired path.

\begin{figure}
\setlength{\belowcaptionskip}{-6pt}
    \centering
    \includegraphics[width=1.0\linewidth]{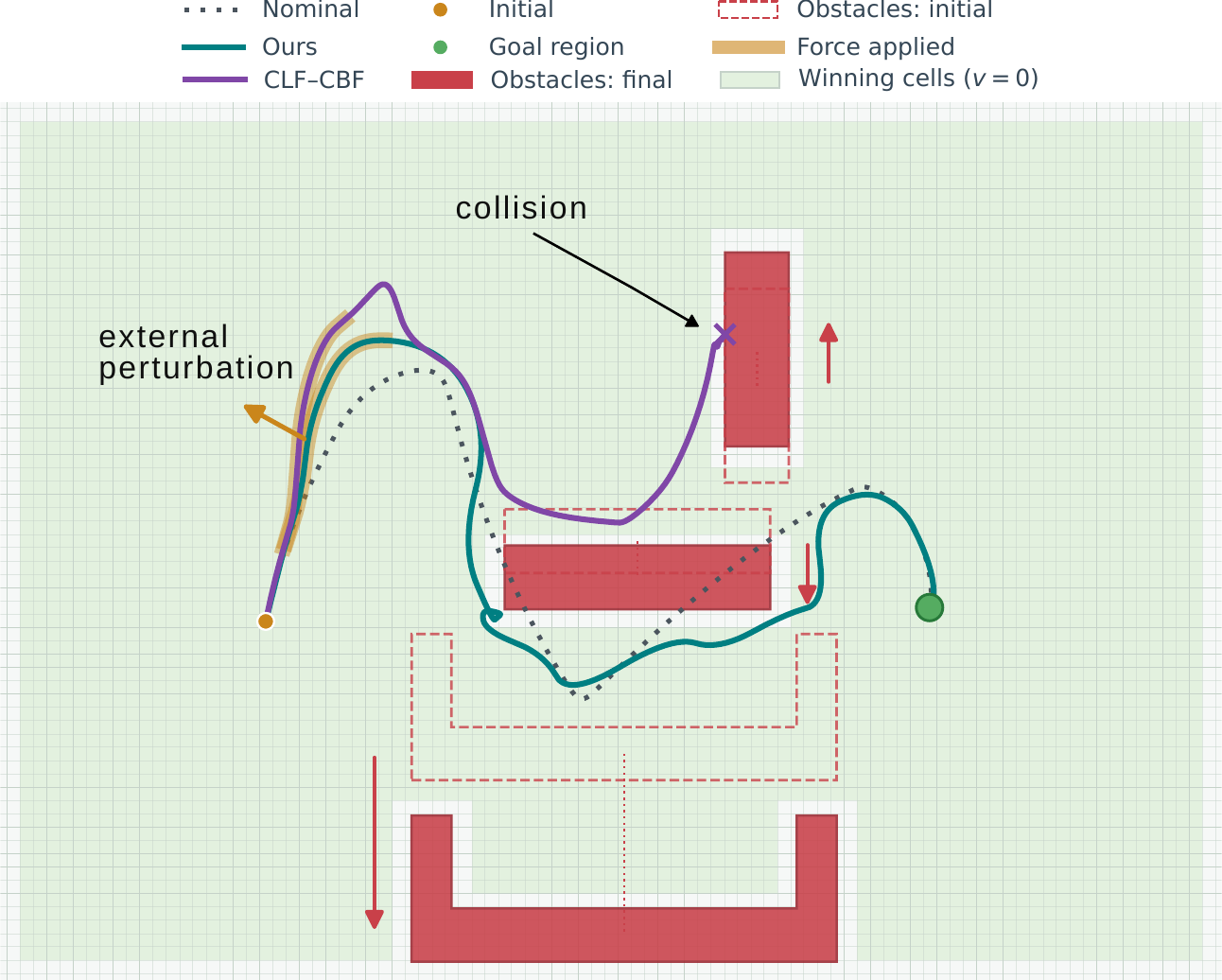}
    \caption{ \centering Online realization of our approach vs CLF-CBF on dynamic obstacles with sine motion; red arrows indicate the motion of the obstacles.}
    \label{fig:online_sim}
\end{figure}

Next, we test our method in the \textit{online} case, where the same learned DS are tested against dynamic obstacles. We use pFaces~\cite{pfaces} as an online synthesis server, which computes at each replanning step (operating at $5Hz$), the robot environment, state, goal set, the obstacle-derived safe set, and the winning set and corresponding backup controller, held and applied at the runtime frequency of $1kHz$. The new velocity and acceleration constraints are $v \in [-0.4, 0.4]$,/s and $u \in [-2,2]$m/s$^2$, respectively. The simulation in Figure~\ref{fig:online_sim} shows the ability of our supervisory controller to to navigate in complex environments with multiple obstacles. Even under perturbations (through the application of external forces), our controller is designed to bring the robot to the intended path and reach the goal in a non-conservative manner. On the other hand, CLF-CBF fails due to collisions at runtime. 


\noindent \textbf{Benchmark Comparisons with CLF-CBF. } For the quantitative evaluation, we sample $10$ DS trajectories from each learning method, generated from $10$ randomly sampled initial positions. This results in a total of $300 = 10 \times 10 \times 3$ motion trials. Among these cases, $17$ are nominal motions without any perturbations or obstacles, $185$ are with obstacles only, $66$ are with perturbations only, and $32$ are realized with both obstacles and perturbations. We evaluate a selected set of performance criteria, including task success rates, safety violations, input constraint violations, and path return rates in the event of trajectory deviations, reported in Table~\ref{tab:performance-rates}, while the break down of success rates according to the learned method family is reported in Table~\ref{tab:outcomes}. All the obstacles considered here are static, and the \textit{supervisory control} is implemented in an offline manner. Note that for CLF-CBF, we use the same tuning parameters provided in~\cite{nawaz}. A task is said to be successful when the robot reaches the goal region within a prescribed time limit without collisions. State constraint violations ( position and velocity), workspace limit violations, task timeout (e.g. due to lack of feasible control inputs), and joint torque limit violations contribute to task failure. We observe that while task failure in our approach is mainly due to timeout and joint limit violations, CLF-CBF is vastly affected by collisions and velocity limit violations as well. Moreover, while our controller is designed to ensure higher path return rates, such criteria isn't considered in CLF-CBF, contributing to lower path return possibilities. Finally, the results demonstrate the formal validity guarantees obtained through our symbolic supervisory control, with no constraint violations (across position, velocity, and input acceleration\footnote{Accelerations for CLF-CBF are computed as $a_k=(v_k-v_{k-1})/\Delta t$, since the QP-based controller does not expose an explicit acceleration command.}) reported across all $300$ trials. 

In addition, we also compare task fidelity criteria, with results reported in Table~\ref{tab:trajectory_metrics}. We observe that our supervisory control approach provides closer imitation to the learned paths, as demonstrated by the lower mean-square-error (MSE) and dynamic-time-warping (DTW) compared to CLF-CBF. In addition, navigation time is also reported to be less in our case, primarily due to the lack of conservatism caused due to CBFs during collision avoidance. However, our method reports a larger acceleration variation among the realizations, leading to lower trajectory smoothness. We posit that this is due to the discretization of the acceleration space for control input computation. 

\begin{table*}[t]
\centering
\caption{Quantitative comparison on the task performance rates between our method and CLF-CBF. All criteria are evaluated over $300$ aggregated DS-learning and tracking tasks, and are measured in $\%$. }
\label{tab:performance-rates}
\resizebox{0.7\textwidth}{!}{
\begin{tabular}{lccccc}
\toprule
\textbf{method} & \textbf{task success} & \textbf{collisions} & \textbf{vel. limit violations} & \textbf{acc. limit violations} & \textbf{path return rate} \\
\midrule
ours (Certified Supervisor) & \textbf{95.67} & \textbf{0.0} & \textbf{0.0} & \textbf{0.0} & \textbf{93.33} \\
CLF-CBF & 65.67 & 18.0 & \textbf{0.0} & 43.33 & 53.33 \\
\bottomrule
\end{tabular}}
\end{table*}

\begin{table}
\centering
\caption{Trajectory-fidelity and motion-quality comparison between the our method and the CLF-CBF baseline. All values are averaged over successful trials, aggregated across $10$ LASA datsets, 
$3$ learned \ac{ds} models per motion (\ac{lpv}-DS, PolyDS, and NN-DS), and $10$ motion trajectories per model.} 
\label{tab:trajectory_metrics}
\resizebox{0.5\textwidth}{!}{
\begin{tabular}{lcccc}
\toprule
\textbf{Method} & \textbf{MSE(mm)} & \textbf{DTW(mm)} & \textbf{acc. variation (m/s$^2$)} & \textbf{navigation time (s)} \\
\midrule
Ours (Certified Supervisor) & \textbf{7.46} & \textbf{14.613} & 0.249 & \textbf{23.27} \\
CLF-CBF& 18.98 & 17.551 & \textbf{0.02} & 24.5\\
\bottomrule
\end{tabular}}
\end{table}

\begin{table}[t]
\centering
\caption{The learning method-family wise breakdown of the task success rates (ours vs CLF-CBF~\cite{nawaz}). }
\label{tab:outcomes}
\small
\resizebox{0.7\columnwidth}{!}{
\begin{tabular}{@{}lrrr@{}}
\toprule
Learning Method & Trials & ours & CBF-CLF \\
\midrule
LPV-DS
& 100
& $99.00\%$
& $54.00\%$ \\

PolyDS
& 100
& $100.00\%$
& $76.00\%$ \\

NNDS
& 100
& $88.00\%$
& $67.00\%$ \\
\bottomrule
\end{tabular}}
\end{table}
\noindent \textbf{Robot Experiments.}
\begin{figure}
    \centering
    \includegraphics[width=0.9\linewidth]{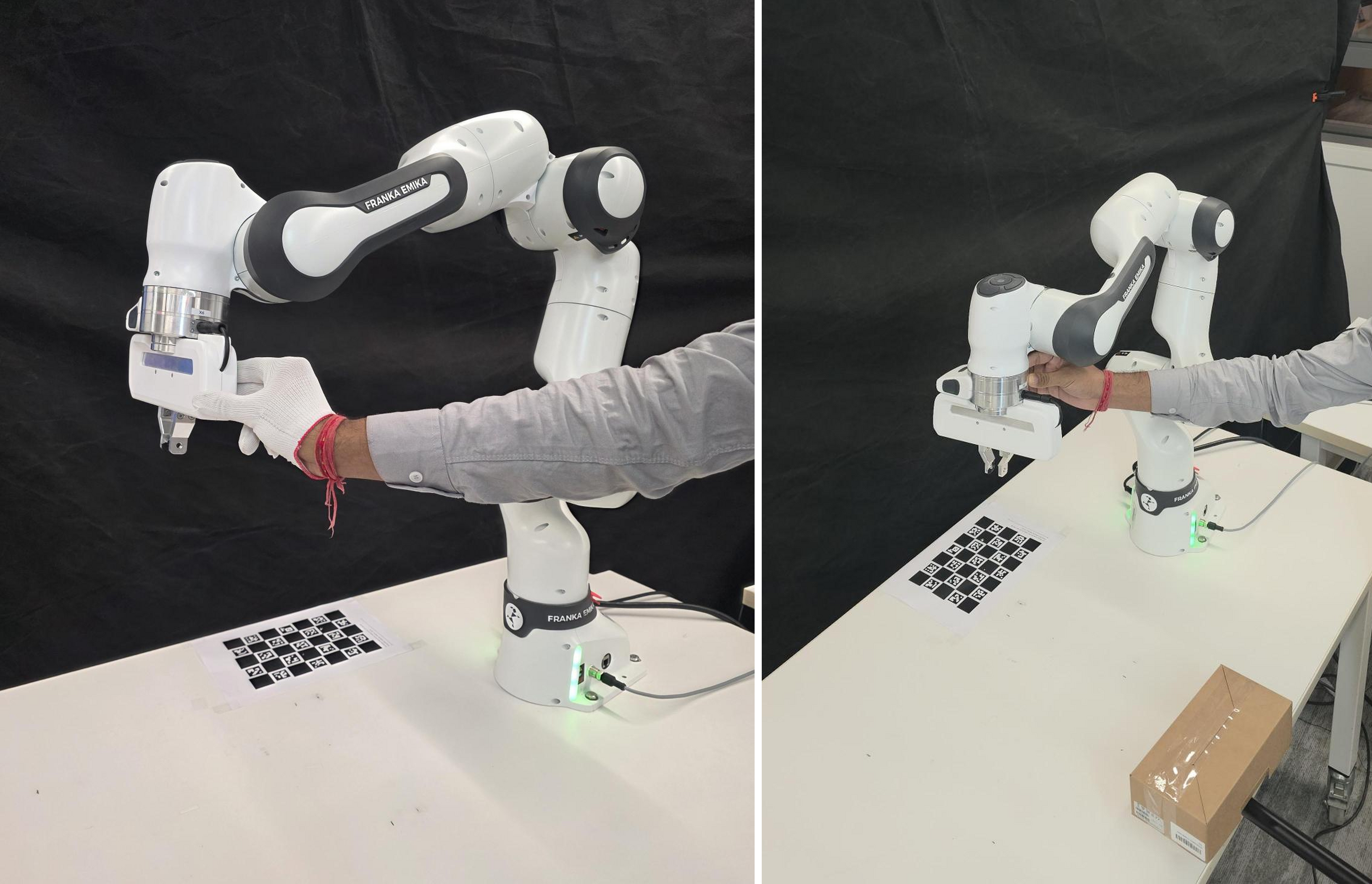}
    \caption{Franka FR3 robot executing a learned C-shaped motion under manually applied perturbations, without an obstacle (left) and with a moving obstacle (right)}
    \label{fig:exp1}
\end{figure}
\begin{figure}
    \centering
    \includegraphics[width=1.0\linewidth]{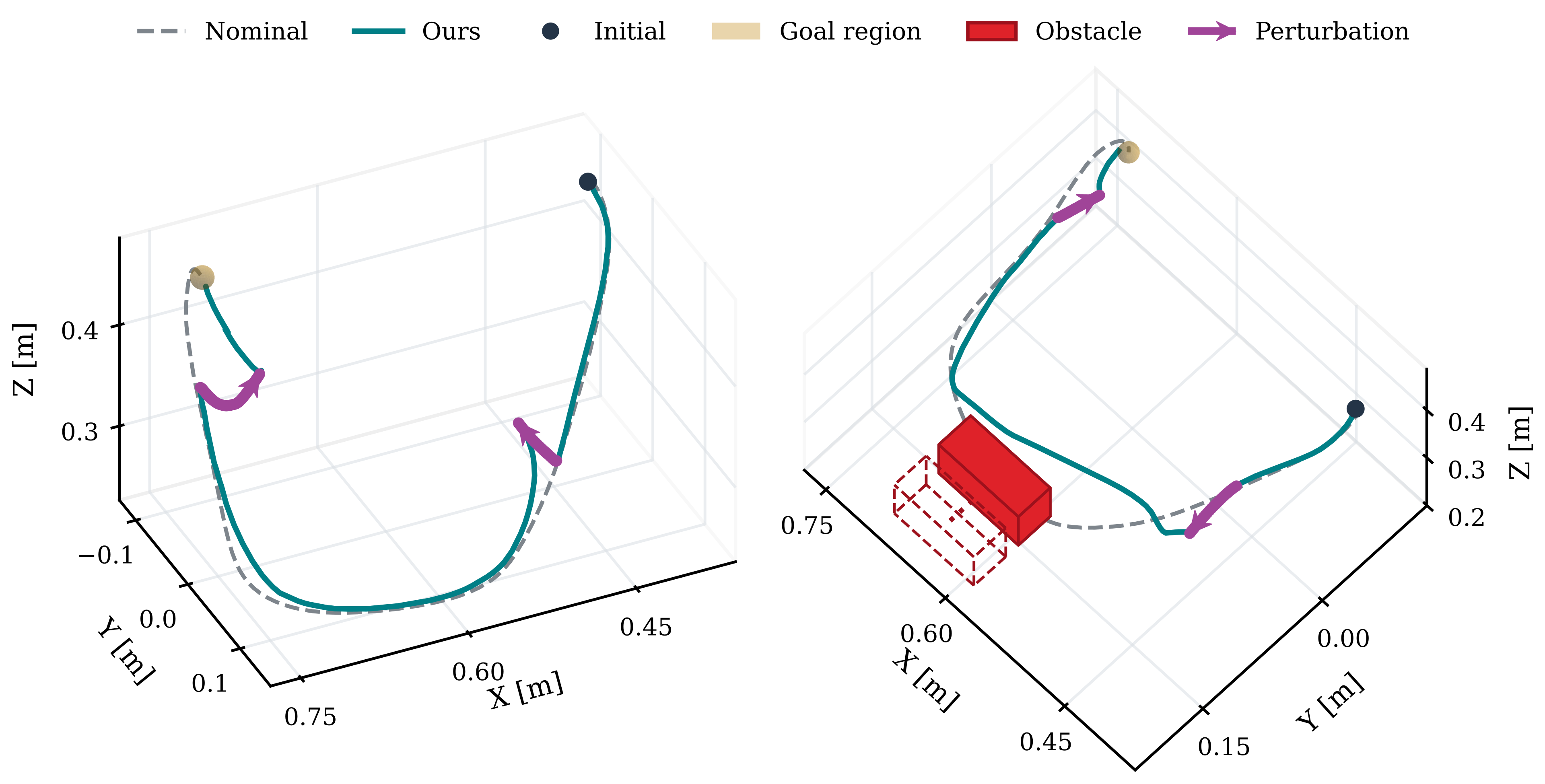}
    \caption{End-effector trajectories when subject to perturbations at runtime using offline (left) and online (right) symbolic synthesis.} 
    \label{fig:exp2}
\end{figure}
We evaluated the proposed framework on a physical Franka FR3 manipulator (see Figure~\ref{fig:exp1}) executing a 3D C-shaped motion\footnote{Available at \url{https://github.com/nbfigueroa/ds-opt}} under offline and online policy execution. Symbolic actions were selected at $5$Hz, corresponding to a $200$ms control interval, with a $1$KHz torque controller for tracking. We provide a constant end-effector orientation tracking reference. The~\ac{ds} was pre-computed using NNDS~\cite{s2nnds} offline before executing the trajectories. The experiments in Figure~\ref{fig:exp2} show the smooth operation of our approach in real time. As perturbations are applied (both in the \textit{offline} and \textit{online} case), the robot demonstrates quick recovery from the perturbed path to the learned one. In the \emph{online} case which additionally incorporated an artificially introduced virtual obstacle, the robot shows quick, reactive collision avoidance through non-conservative path modifications while maintaining navigation toward the goal until it recovers back to the desired path. Note that here, a background synthesis process via pFaces~\cite{pfaces} refreshed the policy using predictions of the virtual obstacle geometry, with update requests issued every $400$ms. The representative offline and online execution, shown in Fig.~\ref{fig:exp2}, reached and held the goal with a final position error of $(8.83, 9.14, 9.57,9.23)$mm from the goal, within the tolerance limit pf $d_\mathrm{tol} = 10$mm in all directions~\eqref{eq:goal_set}. These experiments highlight efficient, reactive execution of the learned DS, and perturbation recovery with both precomputed and online-updated policies, together with obstacle avoidance in the online configuration. 

\section{Conclusion}

We presented a supervisory control methodology that verifies and modifies a nominal controller aimed to track learned \ac{ds}. Our approach is agnostic to the \ac{ds} learning method used, and can generate formally correct actions under mild assumptions on the system and environment. The results are promising, showing competitive performance against SoA. However, we have observed that our method occasionally causes joint limit violations since the low-dimensional task-space abstraction does not account for joint space dynamics. Our future work will aim to extend symbolic certified controller synthesis to higher-dimensional joint spaces through scalable, compositional approaches. 
\section*{Acknowledgements}
This work has been partially supported by the German Federal Ministry of Research, Technology and Space (BMFTR) under the Robotics Institute Germany (RIG).
\balance
\bibliographystyle{IEEEtran}
\bibliography{references}
\end{document}